\documentclass[10pt,twocolumn,letterpaper]{article}

\usepackage[pagenumbers]{cvpr} 

\usepackage{url}
\usepackage{booktabs}       
\usepackage{xspace}
\usepackage{multirow}
\usepackage{makecell}
\usepackage{tablefootnote}
\usepackage{graphicx}
\usepackage{wrapfig}
\usepackage{lipsum}
\usepackage{boxedminipage}
\usepackage{listings}
\usepackage{booktabs}
\usepackage{caption}
\usepackage{colortbl}

\newcommand{\cxmark}{\ding{52}\rotatebox[origin=c]{-9.2}{\kern-0.7em\ding{55}}}

\newcommand{\header}[1]{\text{#1}}

\usepackage[dvipsnames]{xcolor}

\definecolor{Slide}{HTML}{F27970}
\definecolor{Research Report}{HTML}{BB9727}
\definecolor{Financial Report}{HTML}{54B345}
\definecolor{Brochure}{HTML}{32B897}
\definecolor{Academic Paper}{HTML}{FFA500}
\definecolor{Guideline}{HTML}{05B9E2}
\definecolor{Webpage Screenshot}{HTML}{8983BF}
\definecolor{Poster}{HTML}{C76DA2}
\definecolor{Industry File}{HTML}{000000}
\definecolor{darkgreen}{RGB}{50,100,0}
\definecolor{darkred}{RGB}{200, 0, 0}
\definecolor{firstBest}{rgb}{0.86, 1, 0.86}
\definecolor{secondBest}{rgb}{1, 0.91, 0.93}

\definecolor{color5}{HTML}{006795}

\newcommand{\method}{\textsc{M3DocRAG}}
\newcommand{\dataset}{\textsc{M3DocVQA}}

\newcommand{\colpali}{ColPali}
\newcommand{\colqwen}{ColQwen}

\newcommand{\mmqa}{MultimodalQA}

\DeclareMathOperator\argtopk{argtop-k}

\definecolor{cvprblue}{rgb}{0.21,0.49,0.74}
\usepackage[pagebackref,breaklinks,colorlinks]{hyperref}

\hypersetup{colorlinks,linkcolor={color5},citecolor={color5},urlcolor={magenta}}  

\usepackage[capitalize]{cleveref}
\crefname{section}{Sec.}{Secs.}
\Crefname{section}{Section}{Sections}
\Crefname{table}{Table}{Tables}
\crefname{table}{Tab.}{Tabs.}

\def\paperID{*****} 
\def\confName{CVPR}
\def\confYear{2025}

\title{\method{}: Multi-modal Retrieval is What You Need \\for Multi-page Multi-document Understanding}

\author{Jaemin Cho$^1$\thanks{\it{Work done during an internship at Bloomberg as a recipient of the Bloomberg Data Science Ph.D. Fellowship.}}
\quad
Debanjan Mahata$^2$
\quad
Ozan \.Irsoy$^2$
\quad
Yujie He$^2$
\quad
Mohit Bansal$^1$ \\
$^1$UNC Chapel Hill \quad $^2$Bloomberg \\
\texttt{\{jmincho,mbansal\}@cs.unc.edu} \quad
\texttt{\{dmahata,oirsoy,yhe247\}@bloomberg.net}
}

\begin{document}

\maketitle

\begin{abstract}
Document visual question answering (DocVQA) pipelines that answer questions from documents have broad applications.
Existing methods focus on
handling single-page documents with multi-modal language models (MLMs),
or 
rely on
text-based retrieval-augmented generation (RAG)
that uses text extraction tools such as optical character recognition (OCR).
However, there are difficulties in applying these methods in real-world scenarios:
(a) questions often require information across different pages or documents, where MLMs cannot handle many long documents;
(b) documents often have important information in visual elements
such as figures,
but
text extraction tools
ignore them.
We introduce \method{},
a novel multi-modal RAG framework that
flexibly accommodates
various document contexts (closed-domain and open-domain),
question hops (single-hop and multi-hop),
and evidence modalities (text, chart, figure, \etc{}).
\method{} finds relevant documents and answers questions using a multi-modal retriever and an MLM, so that it can efficiently handle single or many documents while preserving visual information.
Since previous DocVQA datasets ask questions in the context of a specific document,
we also present \dataset{}, a new benchmark for evaluating open-domain DocVQA over 3,000+ PDF documents with 40,000+ pages.
In three benchmarks (\dataset{}/MMLongBench-Doc/MP-DocVQA),
empirical results show that \method{} with \colpali{} and Qwen2-VL 7B
achieves superior performance than many strong baselines, including state-of-the-art performance in MP-DocVQA.
We provide comprehensive analyses of different indexing, MLMs, and retrieval models.
Lastly, we qualitatively show that \method{}
can successfully handle various scenarios, such as when relevant information exists across multiple pages and when answer evidence only exists in images.
\end{abstract}

\begin{figure*}[t]
    \centering
    \includegraphics[width=.9\linewidth]{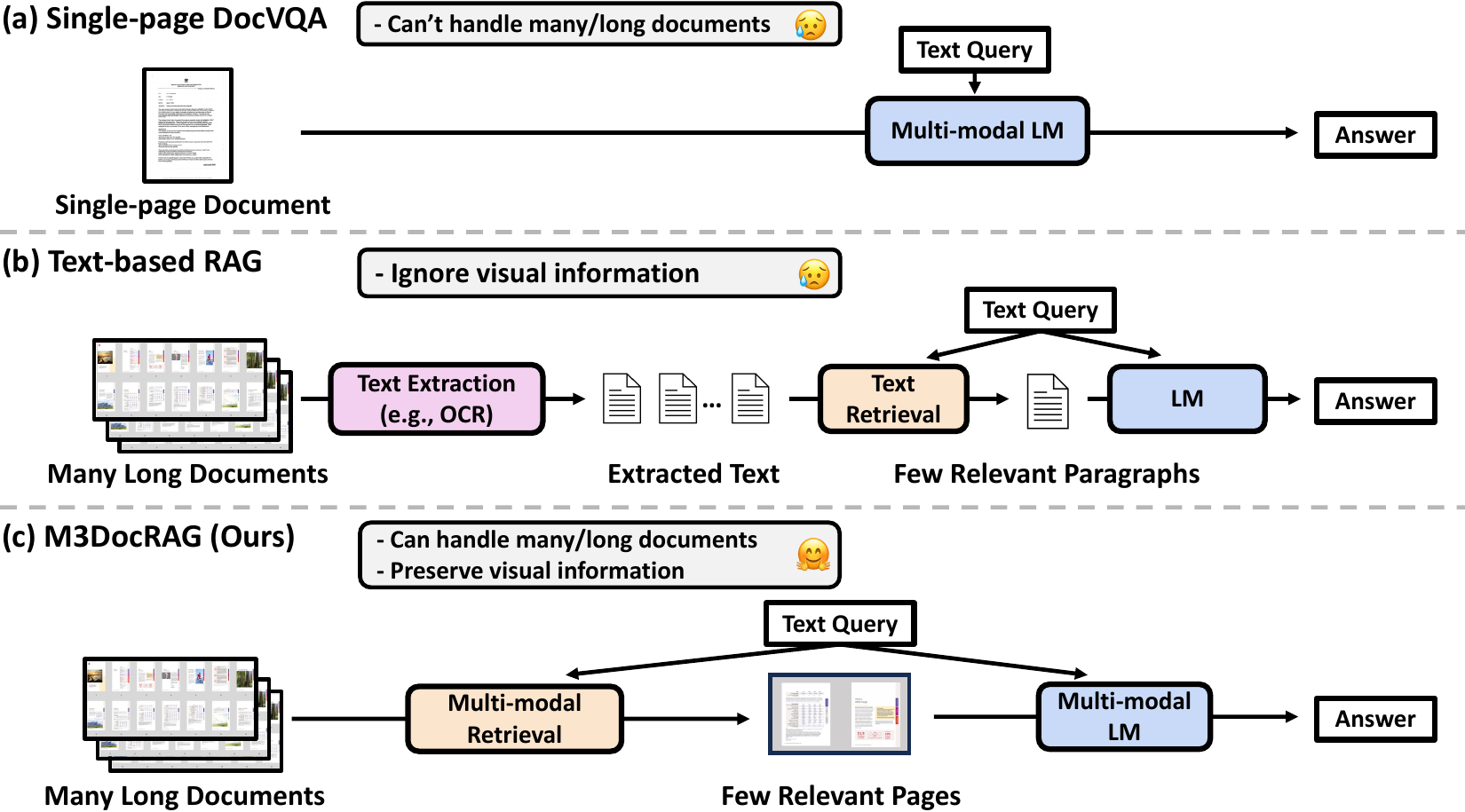}
    \caption{
    Comparison of multi-modal document understanding pipelines.
    Previous works focus on
    \textbf{(a) Single-page DocVQA} that cannot handle many long documents
    or 
    \textbf{(b) Text-based RAG} that ignores visual information.
    Our \textbf{(c) \method{}} framework
    retrieves relevant documents and answers questions using multi-modal retrieval and MLM components, so that it can efficiently handle many long documents while preserving visual information.
    }
    \label{fig:teaser}
\end{figure*}

\section{Introduction and Background}
\label{sec:intro}

Document visual question answering (DocVQA) \citep{mathew2021docvqa,tito2023mpdocvqa,ma2024mmlongbench,ding2023pdfvqa,Landeghem2023dude} is a multi-modal task that answers textual questions by interpreting information contained within document images.
Existing methods on DocVQA
either focus on
visual question answering (VQA) on a single-page document (\cref{fig:teaser} (a))
or 
extract text from documents
(\eg{}, via optical character recognition (OCR)~\citep{smith2007overview,review_ocr}
or PDF text extraction~\citep{pdfminer,pypdf2})
and use retrieval-augmented generation (RAG)~\citep{Lewis2020},
where a retrieval model finds relevant paragraphs and
a language model answers questions given the paragraphs (\cref{fig:teaser} (b)).
However, there are difficulties in applying these methods in real-world document understanding scenarios:
(a) questions often require information across different pages or documents, where existing VQA methods cannot handle many long documents;
(b) some documents feature complex visual formats such as tables, charts, and mixed layouts,
but text extraction methods such as OCR 
ignore these nuances,
leading to incomplete or inaccurate document interpretations. Accurately and efficiently answering questions across numerous, lengthy documents with intricate layouts would greatly benefit many domains such as finance, healthcare, and law, where document AI assistants can streamline the daily processing of large volumes of documents, improving productivity and enabling faster, more informed decision-making.

\begin{figure*}[h]
    \centering
    \includegraphics[width=.75\linewidth]{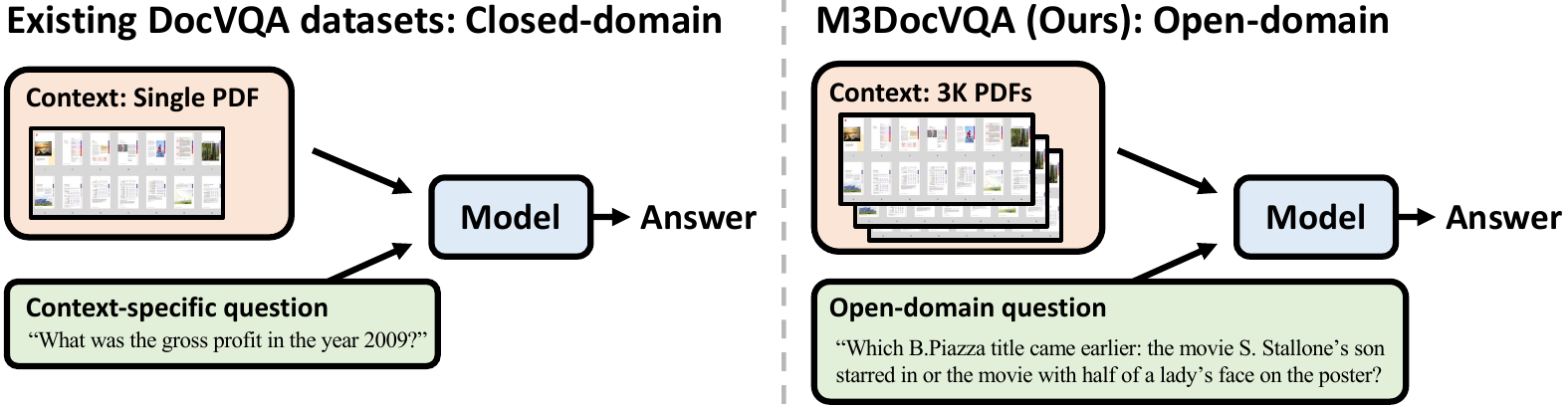}
    \caption{
    Comparison of existing DocVQA datasets (left; \eg{}, DocVQA~\cite{mathew2021docvqa}) and our \textbf{\dataset{}} dataset (right).
    In contrast to previous DocVQA datasets that have questions that are specific to a single provided PDF (\eg{}, ``What was the gross profit in the year 2009?"),
    \dataset{} has information-seeking questions that benchmark open-domain question answering capabilities across more than 3,000 PDF documents (\ie{}, 40,000+ pages).
    }
    \label{fig:dataset}
\end{figure*}

\begin{figure*}[t]
    \centering
    \includegraphics[width=.87\linewidth]{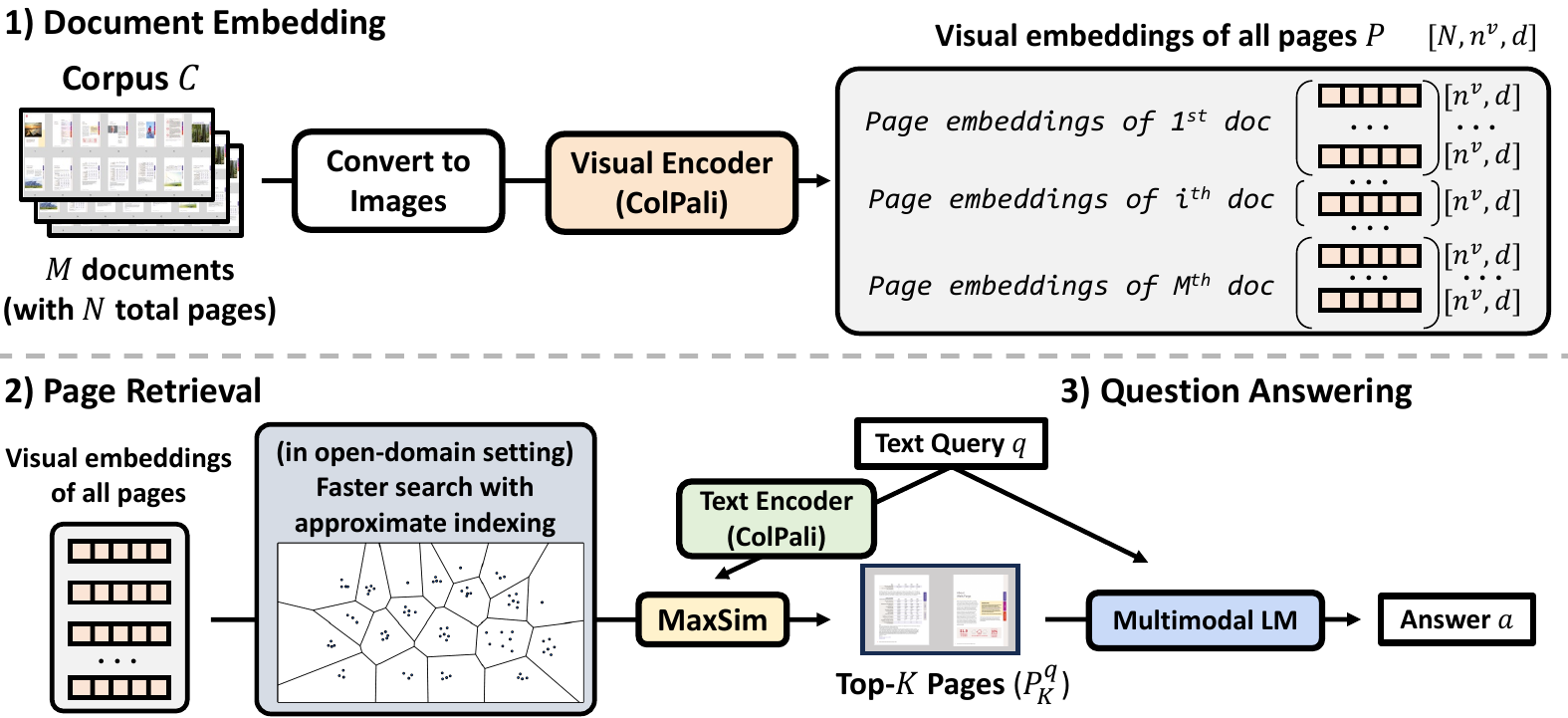}
    \caption{
    Our \method{} framework (\cref{sec:method}) consists of three stages:
    (1) document embedding (\cref{sec:embedding}),
    (2) page retrieval (\cref{sec:retrieval}),
    and (3) question answering (\cref{sec:question_answering}).
    In \textbf{(1) document embedding}, we extract visual embedding (with \colpali{}) to represent each page from all PDF documents.
    In \textbf{(2) page retrieval},  we retrieve the top-K pages of high relevance (MaxSim scores) with text queries.
    In an open-domain setting, we create approximate page indices for faster search.
    In \textbf{(3) question answering}, we conduct visual question answering with multi-modal LM (\eg{} Qwen2-VL) to obtain the final answer.
    }
    \label{fig:method}
\end{figure*}

To overcome these limitations of existing DocVQA approaches,
we
introduce \textbf{\method{}} (\textbf{M}ulti-modal \textbf{M}ulti-page \textbf{M}ulti-\textbf{Doc}ument \textbf{R}etrieval-\textbf{A}ugmented \textbf{G}eneration; \cref{sec:method}),
a novel multi-modal RAG framework that
flexibly accommodates
various document contexts (closed-domain and open-domain),
question hops (single-hop and multi-hop),
and evidence modalities (text, chart, figure, \etc{}).
As illustrated in \cref{fig:teaser} (c), the
\method{} framework
retrieves relevant document pages using a multi-modal retrieval model, such as \colpali{}~\citep{Faysse2024colpali}, and
generates answers to questions from the retrieved pages using a multi-modal language model (MLM), such as Qwen2-VL~\citep{Qwen2VL}.
{\method{}} operates in three stages:
In {(1) document embedding} (\cref{sec:embedding}), we convert all document pages into RGB images and extract visual embeddings (\eg{}, via \colpali{}) from the page images.
In {(2) page retrieval} (\cref{sec:retrieval}), we retrieve the top-K pages of high similarity with text queries (\eg{}, MaxSim operator for \colpali{}).
For the open-domain setting,
we create approximate page indices, such as inverted file index (IVF)~\citep{invertedvideogoogle,invertedfiles}, for faster search.
In {(3) question answering} (\cref{sec:question_answering}), we conduct visual question answering with MLM to obtain the final answer.
Please also see \cref{fig:method} for the detailed illustration of the framework.
\method{} can flexibly handle DocVQA in both closed domain (\ie{}, a single document) and
open-domain (\ie{}, a large corpus of documents) settings.

While \method{} framework supports DocVQA in an open-domain setting,
the existing DocVQA datasets are not adequate for this setting, since their questions are in the context of a specific document, such as ``What was the gross profit in the year
2009?''~\citep{mathew2021docvqa, tito2023mpdocvqa, ding2023pdfvqa, ma2024mmlongbench}, as illustrated in \cref{fig:dataset} (left).
Hence, we also introduce \textbf{\dataset{}} (\textbf{M}ulti-modal \textbf{M}ulti-page \textbf{M}ulti-\textbf{Doc}ument \textbf{V}isual \textbf{Q}uestion \textbf{A}nswering), an open-domain dataset that significantly raises the challenge of DocVQA to answering questions from a large document corpus
(\cref{sec:dataset}).
By extending the MultimodalQA dataset's ~\citep{talmor2021multimodalqa} closed-domain context to an open-domain setting, \dataset{} introduces 2,441 multi-hop questions spanning 3,368 PDF documents, which collectively contain over 41,005 pages of diverse multi-modal content, including text, images, and tables. This dataset presents real-world challenges by requiring models to navigate complex reasoning paths across pages and within various types of document elements, better reflecting the intricacies of document understanding.

To demonstrate the effectiveness of \method{}, 
we compare \method{} with state-of-the-art baselines
in three benchmarks:
\dataset{},
MMLongBench-Doc~\citep{ma2024mmlongbench},
and MP-DocVQA~\citep{tito2023mpdocvqa},
which cover both open-domain (\cref{sec:results_open_domain}) and closed-domain (\cref{sec:results_closed_domain}) DocVQA settings.
Experiment results show that \method{} with \colpali{} and Qwen2-VL 8B
achieves superior performance than many strong baselines,
including the state-of-the-art performance in MP-DocVQA.
We also provide a comprehensive analysis (\cref{sec:ablation}) about different indexing, MLMs, and retrieval components.
Finally, we show qualitative examples (\cref{sec:qual_examples}) where \method{}
can successfully handle various scenarios, such as when the relevant information exists across multiple pages and when answer evidence only exists in images.
Overall, \method{} is an effective, efficient, and flexible framework for answering questions from multi-modal documents in various settings.

\section{\method{}: A Unified Framework for Multi-modal, Multi-page, Multi-document Understanding}
\label{sec:method}

We propose \textbf{\method{}},
a novel multi-modal RAG framework that
flexibly accommodates
various document contexts (closed-domain and open-domain),
question hops (single-hop and multi-hop),
and evidence modalities (text, chart, figure, \etc{}).
As illustrated in \cref{fig:method},
{\method{}} operates in three stages:
(1) encoding document images into visual embeddings (\cref{sec:embedding}),
(2) retrieving relevant document pages (\cref{sec:retrieval}),
and (3) generating answers to questions based on the retrieved pages (\cref{sec:question_answering}).
Below, we explain the problem definition and the details of each stage.

\paragraph{Problem definition.}
We define a corpus of documents as \( C = \{D_1, D_2, \dots, D_M\} \), where \( M \) is the total number of documents, and each document \( D_i \) consists of a set of pages, $P_i$, represented as RGB images. From the documents in \( C \), we construct a global set of page images \( P = \bigcup_{i=1}^M P_i = \{p_1, p_2, \dots, p_N\} \), where each \( p_j \) represents an individual page image, and \( N \) is the total number of page images across all documents in \( C \) (\ie{}, \( N = \sum_{i=1}^{M} |P_i| \)). The objective of {\method{}} is to accurately answer a given question \(q\) using the multi-modal information
available in the corpus of documents \( C \).
First,
we
identify 
$P^{q}_{K}$,
the top \( K \) ($\ll N$) pages that are most relevant to answering the query $q$
from the global page set \( P \).
Then, we obtain the final answer with a question answering model that takes retrieved page images \( P^{q}_K \) and query \( q \) as inputs.
The problem of question answering can be categorized into two  settings with different document context sizes:

\textit{\textbf{Closed-domain question answering}} -- The query \( q \) should be answerable from a given single document \( D_i \). The retrieval model outputs the top \( K \) relevant page images \( P^q_K \), from the page images \( P_i\) of the document \( D_i \).

\textit{\textbf{Open-domain question answering}} --
The query \( q \) may require information from single or multiple documents within the entire document corpus \( C \).
The retrieval model outputs the top \( K \) relevant page images \( P^q_K \) from the entire set of page images \( P \).

\subsection{Document Embedding}
\label{sec:embedding}

In \method{}, both textual query \( q \) and page images \( P \) are projected into a shared multi-modal embedding space using \colpali{}~\citep{Faysse2024colpali}.
\colpali{} is a multi-modal retrieval model based on a late interaction mechanism, which encodes the text and image inputs into unified vector representations and retrieves the top \( K \) most relevant images.
\colpali{} adopts both training objective and similarity scoring from ColBERT~\citep{Khattab2020colbert,Santhanam2022colbert2}, which utilizes a shared architecture to encode either textual or visual inputs. In our framework, each page \(p \subseteq P_i\) of a document \(D_{i}\) is treated as a single image with fixed dimensions (width $\times$ height).

From an image of a page, we extract a dense visual embedding \( E^p \in \mathbb{R}^{n^v \times d} \), where \( n^v \) represents the number of visual tokens per page (which remains constant across all pages), and \( d \) denotes the embedding dimension (\eg{}, 128). For a textual query \( q \), we similarly obtain an embedding \( E^q \in \mathbb{R}^{n^q \times d} \), where \( n^q \) is the number of text tokens.

For efficiency, we treat each page of a document independently. This allows us to flatten all pages in the document corpus \( C \) into a single page-level embedding tensor: \( E^{\text{C}} \in \mathbb{R}^{N \times n^v \times d} \), where \( N \) represents the total number of pages in the entire document corpus, \( n^v \) is the number of visual tokens per page, and \( d \) is the embedding dimension. \method{} can flexibly adapt to different retrieval settings, such as a single-page document (\( N = 1 \)), a single document with multiple pages (\eg{} \( N = 100 \)),
and a large corpus of multi-page documents (\eg{} \( N > 1,000\)).

\subsection{Page Retrieval}
\label{sec:retrieval}

The relevance between the query \( q \) and the page \( p \) is computed using the \texttt{MaxSim} score $s(q, p)$:

\[
    s(q, p) = \sum_{i = 1}^{n^q} \max_{j \in [n^v]} E^q_{i,\cdot} \cdot E^{p}_{j,\cdot}
\]
where \( \cdot \) denotes the dot product, and \( E_{i,\cdot} \in \mathbb{R}^d \) denotes the \( i \)-th row (vector) of the embedding matrix \( E \in \mathbb{R}^{n \times d} \).
We then identify 
$P^{q}_{K}$,
the top \( K \) ($\ll N$) pages that
are most relevant to answering the query $q$; \ie{} we search $K$ pages scoring highest $s(q, p)$.
That is,
\[
P^{q}_{K} = \{p^{q}_{1}, p^{q}_{2}, \dots, p^{q}_{K}\} = \argtopk_{p \in P} s(q, p) \label{eq:retrieve}
\]

\paragraph{Approximate indexing for open-domain page retrieval.}
Searching pages over
in a large document corpus
can be time-consuming and computationally expensive.
When a faster search is desired,
we create page indices offline by applying approximate nearest neighborhood search, based on Faiss~\citep{faiss2021,douze2024faiss}.
We use exact search for closed-domain page retrieval and employ
inverted file index (IVF)~\citep{invertedvideogoogle,invertedfiles} (\texttt{IVFFlat} in Faiss) for an open-domain setting, which could reduce page retrieval latency from 20s/query to less than 2s/query when searching across 40K pages.
See \cref{sec:ablation} for a detailed comparison of speed-accuracy tradeoffs across different indexing methods.

\subsection{Question Answering}
\label{sec:question_answering}

We run visual question answering by giving the text query $q$ and retrieved page images $P^{q}_K$ to a multi-modal language model to obtain the final answer.
For this, we employ multi-modal language models (\eg{} Qwen2-VL~\citep{Qwen2VL}) that consist of a visual encoder
$\texttt{Enc}^{\texttt{Vis}}$ and a language model $\texttt{LM}$.
The visual encoder takes $K$-retrieved page images $P^{q}_K$ as inputs and outputs visual embeddings (different from \colpali{} encoder's outputs). The language model takes the visual embeddings and text embeddings of query $q$ as inputs and outputs the final answer $a$ in the autoregressive manner:
\begin{equation*}
    a = \texttt{LM}(\texttt{Enc}^{\texttt{Vis}}(P^{q}_K), q).
\end{equation*}

\begin{figure*}[t]
    \centering
    \includegraphics[width=\linewidth]{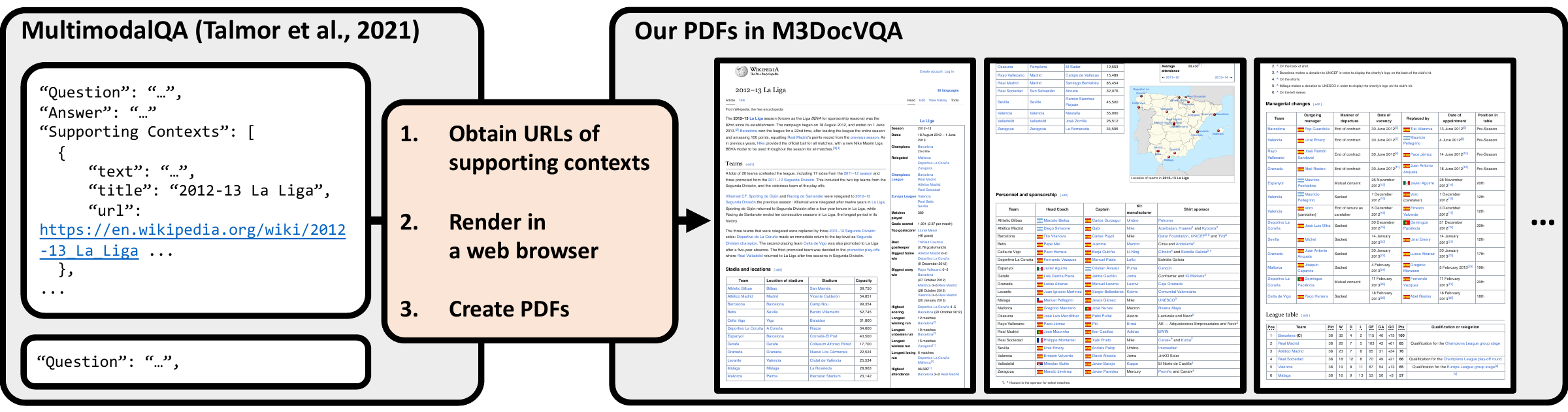}
    \caption{Illustration of PDF collections in \dataset{}.
    We first collect the URLs of all supporting contexts (Wikipedia documents) of individual questions of \mmqa{}~\citep{talmor2021multimodalqa}. Then, we create PDF versions from their URLs by rendering them in a web browser.
    }
    \label{fig:data_collection}
\end{figure*}

\section{\dataset{}: A New Benchmark for Open-domain Document Understanding}
\label{sec:dataset}

We present \textbf{\dataset{}} (\textbf{M}ulti-modal \textbf{M}ulti-page \textbf{M}ulti-\textbf{Doc}ument \textbf{V}isual \textbf{Q}uestion \textbf{A}nswering), a new open-domain DocVQA benchmark designed to evaluate the ability to answer questions using multi-modal information from a large corpus of documents.

As illustrated in \cref{fig:dataset}, existing DocVQA  datasets~\citep{mathew2021docvqa, tito2023mpdocvqa,Landeghem2023dude,ma2024mmlongbench} primarily focus on evaluating question answering within the context of a single document (\ie{}, closed-domain).
These datasets are not well-suited for benchmarking open-domain visual question answering, where relevant information, often in multiple modalities such as text, images, and tables, must be retrieved from multiple documents. This limitation stems from their questions being designed around specific content on certain pages within a single document. 
In real-world scenarios, users often seek answers that span across multiple documents and modalities, making open-domain settings critical.
However, the questions in the existing DocVQA datasets are not applicable in such an open-domain setting.
For example, a question from MP-DocVQA, such as \textit{``What was the gross profit in the year 2009?''} assumes that the model already has access to specific information within the document.

\dataset{} challenges models in an open-domain DocVQA setting,
where they must navigate a large `haystack' of multi-modal documents and retrieve relevant information to generate the final answer.
The dataset consists of 2,441 multi-hop questions spread across 3,368 PDF documents, totaling 41,005 pages.
Each question is supported by evidence found in one or more documents, spanning multiple modalities such as text, images, and tables, capturing the complexity and diversity typical of real-world documents.
Additionally, we provide the training split, consisting of 24,162 Wikipedia PDFs.
Although the documents in the training split were not utilized in our experiments,
they offer future researchers the opportunity to explore even larger-scale retrieval tasks or use the documents for training models, further expanding the potential applications of \dataset{}.

To create \dataset{}, we extend the question-answer pairs from a short-context VQA dataset to a more complex setting that includes 1) PDF documents and 2) open-domain contexts. Specifically, we use the question-answer pairs from the development split\footnote{The test split of \mmqa{}~\citep{talmor2021multimodalqa} is unavailable, and previous works have used the development split for comparison.} of \mmqa{}~\citep{talmor2021multimodalqa}, where models answer multi-hop questions based on short multi-modal contexts (\eg{}, short text passages, 1-2 images, a table) sourced from Wikipedia. We retrieved the URLs of all Wikipedia documents used as context in any of the \mmqa{} development split questions.
Then we generated PDF versions of the Wikipedia pages by rendering them in a Chromium web browser~\citep{chromium}, using the Playwright Python package~\citep{playwright}. These PDFs retain all vector graphics and metadata, ensuring zoom-in functionality and maintaining operational hyperlinks. In addition, no objects are split between different pages in the resulting PDFs.

\begin{table*}[t]
 \centering
 \caption{Open-domain DocVQA evaluation results on \dataset{}.
 The scores are based on F1, unless otherwise noted.
 Index: \texttt{FlatIP} + \texttt{IVFFlat}.
 }
 \label{tab:results_m3docvqa}
 \resizebox{.8\linewidth}{!}{
    \begin{tabular}{l c ccc cc cc}
    \toprule
     \multirow{2}{*}{\textbf{Method}} & \multirow{2}{*}{\textbf{\# Pages}} & \multicolumn{3}{c}{\textbf{Evidence Modalities}} & \multicolumn{2}{c}{\textbf{Question Hops}} & \multicolumn{2}{c}{\textbf{Overall}} \\

     \cmidrule(lr){3-5}
     \cmidrule(lr){6-7}
     \cmidrule(lr){8-9}

     & & \multicolumn{1}{c}{Image} & Table & Text & \multicolumn{1}{c}{Single-hop} & \multicolumn{1}{c}{Multi-hop} & EM & F1 \\

    \midrule
    \textcolor{gray}{\textit{Text RAG (w/ ColBERT v2)}} \\

    Llama 3.1 8B& 1 & 8.3 & 15.7 & 29.6 & 25.3 & 12.3 & 15.4 & 20.0\\

    Llama 3.1 8B& 2 & 7.7 & 16.8 & 31.7 & 27.4 & 12.1 & 15.8 & 21.2\\

    Llama 3.1 8B& 4  & 7.8 & 21.0 & 34.1 & 29.4 & 15.2 & 17.8 & 23.7\\

    \midrule
    \textcolor{gray}{\textit{\method{} (w/ \colpali{})}} \\

    \rowcolor{blue!8} Qwen2-VL 7B (Ours) & 1 & 25.1 & 27.8 & 39.6 & 37.2 & 25.0 & 27.9 & 32.3\\

    \rowcolor{blue!8} Qwen2-VL 7B (Ours) & 2 & \textbf{26.8} & \textbf{30.4} & \textbf{42.1} & 41.0 & 25.2 & 29.9 & 34.6\\

    \rowcolor{blue!8} Qwen2-VL 7B (Ours) & 4 & 24.7 & \textbf{30.4} & 41.2 & \textbf{43.2} & \textbf{26.6} & \textbf{31.4} & \textbf{36.5}\\
    
    \bottomrule
    \end{tabular}
}
\end{table*}

While both \dataset{} and \mmqa{}~\citep{talmor2021multimodalqa} share the goal of evaluating
question answering given multi-modal context, \dataset{} introduces a more demanding scenario by requiring models to retrieve relevant information from a large set of documents, as opposed to being provided with a short context.
In \mmqa{}, models are given short, curated context (\eg{}, a paragraph from a Wikipedia document) that directly contains the information needed to answer the questions, simplifying the task to reasoning within the provided material. In contrast, \dataset{} presents an open-domain setting, where models must retrieve information from a diverse collection of 3,368 PDF documents before attempting to answer any question.
This not only requires handling large-scale document retrieval but also dealing with multi-modal content--text, images, and tables--distributed across multiple documents. This key distinction highlights \dataset{}’s ability to simulate real-world challenges, where the relevant data is often spread across multiple sources. Consequently, \dataset{} serves as a robust benchmark for retrieval-augmented generation tasks in document understanding, pushing the boundaries of models to deal with large-scale, multi-modal, and multi-document settings.

\section{Experiment Setup}
\label{sec:exp_setup}

\paragraph{Datasets.}
We benchmark \method{} on three PDF document understanding datasets that represent different scenarios:
(1) \dataset{} (Open-domain DocVQA);
(2) MMLongBench-Doc~\citep{ma2024mmlongbench} (Closed-domain DocVQA);
(3) MP-DocVQA~\citep{tito2023mpdocvqa} (Closed-domain DocVQA).
In \dataset{}, \method{} processes over 3,000 PDFs, totaling more than 40,000 pages.
For MP-DocVQA, models handle a single PDF with up to 20 pages for each question.
For MMLongBench-Doc, models handle a single PDF with up to 120 pages for each question.

\paragraph{Evaluation Metrics.}
For \dataset{}, we follow the evaluation setup of \mmqa{}~\citep{talmor2021multimodalqa}.
For MMLongBench-Doc~\citep{ma2024mmlongbench} and MP-DocVQA~\citep{tito2023mpdocvqa}, we follow their official evaluation setups.
For \dataset{}, we evaluate answer accuracy with exact match (EM) and F1.
For MMLongBench-Doc,
we extract short answers with GPT4o~\citep{gpt4o} from the model outputs and 
report answer accuracy with generalized accuracy (based on a rule-based evaluation script covering different answer types) and F1 score.
For MP-DocVQA, we report answer accuracy with ANLS~\citep{Biten2019} and page retrieval with accuracy (same as recall@1, as there is a single page annotation for each question) by submitting the generation results to the test server.\footnote{\url{https://rrc.cvc.uab.es/?ch=17&com=tasks}}

\paragraph{Models.}
We mainly experiment with the \colpali{} v1~\citep{Faysse2024colpali}\footnote{\url{https://huggingface.co/vidore/colpali}} retrieval model and various recent open source multi-modal LMs with $<$10B parameters,
including Idefics 2~\citep{laurencon2024mattersidefics2}, Idefics 3~\citep{laurençon2024idefics3}, InternVL 2~\citep{chen2024internvl2}, and Qwen2-VL~\citep{Qwen2VL}.
We also experiment with a text-based RAG pipeline by combining recent widely used text retrieval and language models:
ColBERT v2~\citep{Santhanam2022colbert2} and Llama 3.1~\citep{dubey2024llama3herdmodels}.
We also compare \colpali{} v1 with \colqwen{} v0.1~\citep{Faysse2024colpali},\footnote{\url{https://huggingface.co/vidore/colqwen2-v0.1}} another recent multi-modal retrieval model that was trained with same objective/dataset as \colpali{} but initialized with Qwen2-VL 2B~\citep{Qwen2VL} backbone.
For reproducible evaluation, we use deterministic greedy decoding for answer generation.
We compare these multi-modal and text-based RAG pipelines with recent top entries with comparable parameters ($<$10B) reported on the leaderboards.

\paragraph{Other implementation details.}
We use PyTorch~\citep{Paszke2017,Paszke2019}, Transformers~\citep{Wolf2019}, and FlashAttention-2~\citep{dao2023flashattention2} libraries for running models.
We use Tesseract~\citep{smith2007overview} for OCR in text RAG baselines, following \citet{ma2024mmlongbench}.
We use Faiss~\citep{faiss2021,douze2024faiss} for document indexing.
We use the pdf2image~\citep{pdf2image} library to convert each PDF page into an RGB image with a resolution of DPI=144.
While all PDF pages in \dataset{} have the same size -- 8.5 (width) $\times$ 11 (height) in inches (\ie{} US letter size) and 1224 (width) $\times$ 1584 (height) in pixels,
in MP-DocVQA and MMLongBench-Doc datasets, pages have slightly different sizes.
To handle this, we resize page images to the most common image size within the dataset -- 1700 (width) $\times$ 2200 (height) for MP-DocVQA,
and to the most common image size within each PDF document for MMLongBench-Doc.
All experiments are conducted with a single H100 80GB GPU.
We provide up to 4 pages as visual inputs to our multi-modal LMs, the maximum number of images we could fit in the single GPU.

\begin{table*}[t]
 \centering
 \footnotesize
 \caption{Closed-domain DocVQA evaluation results on MMLongBench-Doc.
 We report the generalized accuracy (ACC) across five evidence source modalities: text (TXT), layout (LAY), chart (CHA), table (TAB), and image (IMG), and three evidence locations: single-page (SIN), cross-page (MUL), and unanswerable (UNA).
 The scores from non-RAG methods are from \citet{ma2024mmlongbench}.
 }
 \label{tab:results_mmlongbench_doc}
 \resizebox{.85\linewidth}{!}{
    \begin{tabular}{l c |ccccc|ccc|cc}
    \toprule
     \multirow{2}{*}{\textbf{Method}} & \multirow{2}{*}{\textbf{\# Pages}} &  \multicolumn{5}{c|}{\textbf{Evidence Modalities}} & \multicolumn{3}{c|}{\textbf{Evidence Locations}} & \multicolumn{2}{c}{\textbf{Overall}}  \\

    \cmidrule(lr){3-7}
    \cmidrule(lr){8-10}
    \cmidrule(lr){11-12}
     
     & & \header{TXT} & \header{LAY} & \header{CHA} & \header{TAB} & \header{IMG} & SIN & MUL & UNA & \multirow{1}{*}{ACC} & \multirow{1}{*}{F1}\\ 
    \midrule
    \multicolumn{12}{l}{\hfill \textit{Text Pipeline} } \\
    \midrule
    \textcolor{gray}{\textit{LMs}} \\
    ChatGLM-128k~\citep{bai2024longalign}  & up to 120   & 23.4 & 12.7 & 9.7 & 10.2 & {12.2} & 18.8 & 11.5 & 18.1 & 16.3 & 14.9 \\
    Mistral-Instruct-v0.2~\citep{jiang2023mistral} & up to 120 & 19.9 & 13.4 & 10.2 & 10.1 & 11.0 & 16.9 & 11.3 & 24.1 & 16.4 & 13.8 \\
    \textcolor{gray}{ \textit{Text RAG}} \\
    ColBERT v2 + Llama 3.1 & 1 & 20.1 & 14.8 & 12.7 & 17.4 & 7.4 & 21.8 & 7.8 & \textbf{41.3} & 21.0 & 16.1 \\
    ColBERT v2 + Llama 3.1 & 4  & {23.7} & {17.7} & {14.9} & \textbf{24.0} & 11.9 & 25.7 & 12.2 & 38.1 & \textbf{23.5} & {19.7} \\

    \midrule

    \multicolumn{12}{l}{\hfill \textit{Multi-modal Pipeline}} \\
    \midrule
    \textcolor{gray}{\textit{Multi-modal LMs}} \\
    DeepSeek-VL-Chat~\citep{lu2024deepseek} & up to 120& 7.2 & 6.5 & 1.6 & 5.2 & 7.6 & 5.2 & 7.0 & \textbf{12.8} & 7.4 & 5.4 \\
    Idefics2~\citep{laurencon2024mattersidefics2} & up to 120 & 9.0 & 10.6 & 4.8 & 4.1 & 8.7 & 7.7 & 7.2 & 5.0 & 7.0 & 6.8 \\
    MiniCPM-Llama3-V2.5~\citep{yu2024rlaifv,xu2024llava-uhd} & up to 120 & 11.9 & 10.8 & 5.1 & 5.9 & 12.2 & 9.5 & 9.5 & 4.5 & 8.5 & 8.6 \\
    InternLM-XC2-4KHD~\citep{dong2024internlm} & up to 120  & 9.9 & 14.3 & 7.7 & 6.3 & 13.0 & 12.6 & 7.6 & 9.6 & 10.3 & 9.8 \\
    mPLUG-DocOwl 1.5~\citep{hu2024mplugdocowl} & up to 120 & 8.2 & 8.4 & 2.0 & 3.4 & 9.9 & 7.4 & 6.4 & 6.2 & 6.9 & 6.3 \\
    Qwen-VL-Chat~\citep{bai2023qwenvl} & up to 120 & 5.5 & 9.0 & 5.4 & 2.2 & 6.9 & 5.2 & 7.1 & 6.2 & 6.1 & 5.4 \\
    Monkey-Chat~\citep{li2023monkey} & up to 120 & 6.8 & 7.2 & 3.6 & 6.7 & 9.4 & 6.6 & 6.2 & 6.2 & 6.2 & 5.6 \\
    
    \textcolor{gray}{\textit{\method{}}} \\

    \rowcolor{blue!8} \colpali{} + Idefics2  (Ours) & 1 & 10.9 & 11.1 & 6.0 & 7.7 & 15.7 & 15.4 & 7.2 & 8.1 & 11.2 & 11.0 \\

    \rowcolor{blue!8} \colpali{} + Qwen2-VL 7B (Ours) & 1 & 25.7 & 21.0 & 18.5 & 16.4 & 19.7 & 30.4 & 10.6 & 5.8 & 18.8 & 20.1 \\
    
    \rowcolor{blue!8} \colpali{} + Qwen2-VL 7B (Ours) & 4 & \textbf{30.0} & \textbf{23.5} & \textbf{18.9} & {20.1} & \textbf{20.8} & \textbf{32.4} & \textbf{14.8} & 5.8 & {21.0} & \textbf{22.6} \\
    
    \bottomrule
    \end{tabular}
    }
\end{table*}

\section{Results and Key Findings}
\label{sec:results}

In the following,
we describe experiment results of \method{} and baselines in both open-domain (\cref{sec:results_open_domain}) and closed-domain settings (\cref{sec:results_closed_domain}).
Next, we provide ablation studies (\cref{sec:ablation}) about different page indexing strategies and
different multi-modal LMs and retrieval models.
Lastly, we show qualitative examples (\cref{sec:qual_examples}) where \method{} can tackle \dataset{} questions whose answer source exists in various modalities.

\subsection{Open-domain DocVQA}
\label{sec:results_open_domain}

\paragraph{Multi-modal RAG outperforms text RAG, especially on non-text evidence sources.}
\Cref{tab:results_m3docvqa} shows the evaluation results on \dataset{}.
As a model needs to find relevant documents from 3,000+ PDFs for each question, we focus solely on RAG pipelines.
We observe that our \method{} (\colpali{} + Qwen2-VL 7B) significantly outperforms text RAG (ColBERT v2 + Llama 3.1 8B), across all different evidence modalities / question hops / \# pages.
The performance gap is especially big when the evidence involves images, underscoring that \method{} addresses the information loss over non-textual content by text-only pipelines.
We also notice that providing more retrieved pages as context generally increases the performance of both text RAG and \method{} (using the top 4 pages gives higher performance than the top 1 and 2 pages).

\subsection{Closed-domain DocVQA}
\label{sec:results_closed_domain}

\paragraph{Multi-modal RAG boosts long document understanding of MLMs.}
In MMLongBench-Doc, the models must handle a long PDF document (up to 120 pages) for each question.
Since many multi-modal LMs have limited context length, \citet{ma2024mmlongbench} employed a concatenation strategy that combines all screenshot pages into either 1 or 5 images and inputs these concatenated images to multi-modal LMs.
\Cref{tab:results_mmlongbench_doc} shows that
\colpali{} + Idefics2 surpass 
Idefics2 without RAG, as well as all previous multi-modal entries.
In addition, \colpali{} + Qwen2-VL 7B achieves the best scores in overall F1 and most evidence modality/page settings. 
This demonstrates the effectiveness of multi-modal retrieval over handling many pages by concatenating low-resolution images.
As observed in \dataset{} experiments,
we also notice that providing more retrieved pages as context generally increases the performance of both text RAG and \method{} (using the top 4 pages gives higher performance than the top 1 page).

\paragraph{\method{} achieves the state-of-the-art performance in MP-DocVQA.}
In MP-DocVQA, the models must handle a PDF document of up to 20 pages for each question.
\Cref{tab:results_mp_docvqa} presents the top-performing entries in the MP-DocVQA test split leaderboard, comparing text-based and multi-modal RAG pipelines. While the text RAG (ColBERT v2 + Llama 3.1) falls short compared to existing approaches, all multi-modal RAG pipelines outperform their text-based counterpart.
Notably, the \method{} pipeline (\colpali{} + Qwen2-VL 7B) delivers the state-of-the-art results on MP-DocVQA.
It is interesting that while the existing entries were fine-tuned specifically for MP-DocVQA,
the components used in \method{} (\colpali{} or Qwen2-VL 7B) were not tailored to this dataset -- although Qwen2-VL 7B might have been trained on DocVQA~\citep{mathew2021docvqa}, which shares some images with MP-DocVQA.

\begin{table}[t]
\centering
\caption{Closed-domain DocVQA evaluation results on MP-DocVQA.
The RAG methods retrieve a single page to the downstream QA models.}
\label{tab:results_mp_docvqa}
\resizebox{\linewidth}{!}{
\begin{tabular}{l c c}
\toprule

\multirow{2}{*}{\textbf{Method}}  & \textbf{Answer Accuracy}  & \textbf{Page Retrieval} \\
                                          & {ANLS}              & {R@1}          \\
\midrule
\textcolor{gray}{\textit{Multimodal LMs}}  \\

Arctic-TILT 0.8B~\citep{Borchmann2024arctictilt}          & {0.8122} & {50.79}  \\
GRAM~\citep{blau2024gram}                                & {0.8032} & {19.98}  \\
GRAM C-Former~\citep{blau2024gram}                        & {0.7812} & {19.98}  \\
ScreenAI 5B~\citep{Baechler2024screenai}                    & {0.7711} & {77.88} \\

\midrule

\textcolor{gray}{\textit{Text RAG}} \\
{ColBERT v2 + Llama 3.1 8B}  & 0.5603	& 75.33 \\

\textcolor{gray}{\textit{\method{}}} \\

\rowcolor{blue!8}{\colpali{} + Qwen2-VL 7B} (Ours) & \textbf{0.8444} &	\textbf{81.05}  \\

\bottomrule
\end{tabular}
}
\end{table}
\begin{table}[t]
 \centering
 \caption{Speed-accuracy tradeoff with different indexing strategies on \dataset{}.
 Method: \colpali{} + Qwen2-VL 7B.
 }
 \label{tab:ablation_indexing}
 \resizebox{\linewidth}{!}{
    \begin{tabular}{c c c c c c c}
    \toprule
     \multirow{2}{*}{\textbf{\# Pages}} & \multirow{2}{*}{\textbf{Indexing}} & \multicolumn{2}{c}{\textbf{Latency (s) ($\downarrow$)}}  & \multicolumn{2}{c}{\textbf{Accuracy ($\uparrow$) }} \\

     \cmidrule(lr){3-4}
     \cmidrule(lr){5-6}
     & & Retrieval & {VQA} & {EM} & {F1} \\ 
    \midrule
    1 & FlatIP & 21.0 & 1.1 & 28.9 & 33.7\\
    \rowcolor{blue!8} 1 & FlatIP + IVFFlat & 1.8 & 1.1 &  27.9 & 32.3\\
    1 & FlatIP + IVFPQ & 0.2 & 1.1 & 25.9 & 30.3 \\

    \midrule

    \rowcolor{blue!8} 2 & FlatIP + IVFFlat & 1.8 & 2.4 & 29.9 & 34.6\\
    2 & FlatIP + IVFPQ & 0.2 & 2.4 & 29.0 & 33.5\\

    \midrule

    \rowcolor{blue!8} 4 & FlatIP + IVFFlat & 1.8 & 4.8 & 31.4 & 36.5\\
    4 & FlatIP + IVFPQ & 0.2 & 4.8 & 29.9 & 34.7\\
    
    \bottomrule
    \end{tabular}
    }
\end{table}

\subsection{Additional analysis}
\label{sec:ablation}

\paragraph{Different page indexing: speed and accuracy.}

In \Cref{tab:ablation_indexing}, we analyze the speed and accuracy of \colpali{}+Qwen2-VL 7B pipeline with different document embedding indexing methods.
While the naive indexing with exact search (\texttt{FlatIP}) is slow (21s per query),
we find that using approximate indexing such as inverted file~\citep{invertedfiles,invertedvideogoogle} (\texttt{IVFFlat}) and product quantization~\citep{productquantization} (\texttt{IVFPQ}) can retain most of the accuracy, while making the search significantly faster ($<2$s per query). We use \texttt{FlatIP}+\texttt{IVFFlat} indexing by default, and users can choose appropriate indexing methods depending on their deployment requirements.

\begin{table}[!htbp]
 \centering
 \caption{Comparison of different multimodal LMs within \method{}, evaluated across different document understanding benchmarks.
 For retrieval, we use the top-1 page from \colpali{} for all datasets.
 We use \texttt{FlatIP}+\texttt{IVFFlat} indexing for \dataset{}.
 }
 \label{tab:ablation_mlms}
 \resizebox{\linewidth}{!}{
\begin{tabular}{l ccc}
\toprule

\multirow{2}{*}{\textbf{Multimodal LMs}} &
\textbf{\dataset{}} &
\textbf{MMLongBench-Doc}  &
\textbf{MP-DocVQA}
\\ 

\cmidrule(lr){2-2}
\cmidrule(lr){3-3}
\cmidrule(lr){4-4}

& F1 ($\uparrow$) &  Acc ($\uparrow$) & ANLS ($\uparrow$) \\ 
\midrule
{Idefics2 8B} &  27.8 & 10.8  & 0.56 \\
{Idefics3 8B} & 31.8 & 16.4& 0.77  \\
{InternVL2 8B} & 30.9 & 17.3 & 0.81   \\
\rowcolor{blue!8} {Qwen2-VL 7B} & \textbf{32.3}& \textbf{18.8} &  \textbf{0.84}  \\

\bottomrule
\end{tabular}
}
\end{table}

\paragraph{Different multi-modal LMs.}
In \Cref{tab:ablation_mlms}, we compare four different multi-modal LMs in the \method{} framework: Idefics2 8B~\citep{laurencon2024mattersidefics2}, Idefics3 8B~\citep{laurençon2024idefics3}, InternVL2 8B~\citep{chen2024internvl2}, and Qwen2-VL 7B~\citep{Qwen2VL}.
The Qwen2-VL 7B model outperforms other MLMs in all three benchmarks. Thus, we use the model as our default MLM component.

\begin{table}[!htbp]
 \centering
 \caption{Comparison of different multi-modal retrieval models within \method{} framework, evaluated across different document understanding benchmarks.
 We provide Qwen2-VL 7B with top-4 pages for MMLongBench-Doc/\dataset{} and top-1 page for MP-DocVQA from the retrieval models.
 We use \texttt{FlatIP}+\texttt{IVFFlat} indexing for \dataset{}.
 }
 \label{tab:ablation_retriever}
 \resizebox{\linewidth}{!}{
\begin{tabular}{l ccc}
\toprule

\multirow{2}{*}{\textbf{Ret. Models}} &
\textbf{\dataset{}} &
\textbf{MMLongBench-Doc}  &
\textbf{MP-DocVQA}
 \\ 

\cmidrule(lr){2-2}
\cmidrule(lr){3-3}
\cmidrule(lr){4-4}

&
F1 ($\uparrow$)  &
Acc ($\uparrow$) &
ANLS ($\uparrow$)\\ 
\midrule

\rowcolor{blue!8} {\colpali{} v1} & \textbf{36.5}  & 21.0 & 0.84\\
{\colqwen{} v0.1} & 32.1& \textbf{21.5}  & \textbf{0.86} \\

\bottomrule
\end{tabular}
}
\end{table}

\paragraph{Different multi-modal retrieval models.}
In \Cref{tab:ablation_retriever}, we compare two different multi-modal retrival models in \method{} framework: \colpali{} v1 and \colqwen{} v0.1 (see \cref{sec:exp_setup} for details). Both models are trained with the same training objectives but are initialized with different MLM architectures: PaliGemma 2B~\citep{Beyer2024paligemma} and Qwen2-VL 2B~\citep{Qwen2VL}, respectively.
We find that \colpali{} achieves significantly better performance in \dataset{}, while \colqwen{} achieves slightly better performance in MP-DocVQA and MMLongBench-Doc. Thus, we use \colpali{} as our default retrieval model.

\begin{figure*}[h]
    \centering
    \includegraphics[width=.65\linewidth]{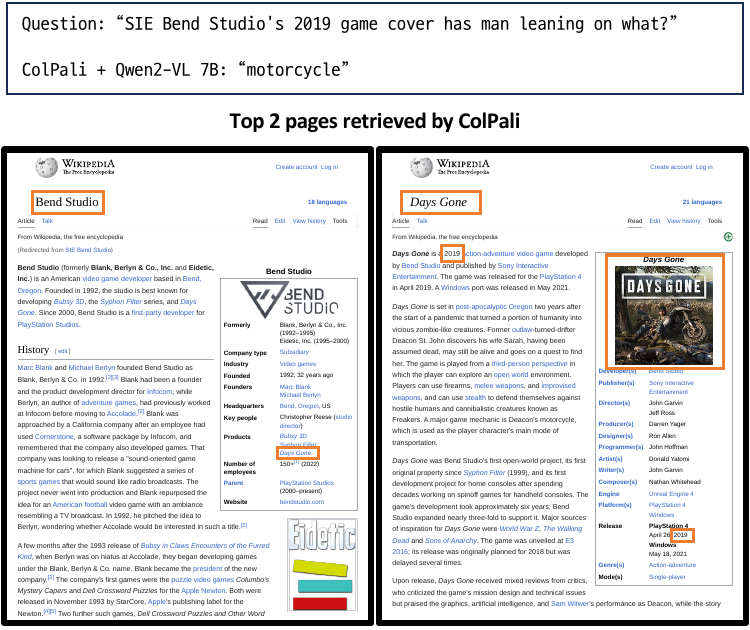}
    \caption{Qualitative example of \colpali{} + Qwen2-VL 7B on \dataset{}. Image regions relevant to the question/answer are highlighted with orange boxes. The answer information is only stored visually within the game logo, where a man is leaning on a motorcycle.}
    \label{fig:qual_example_daysgone}
\end{figure*}
\begin{figure*}[h]
    \centering
    \includegraphics[width=.65\linewidth]{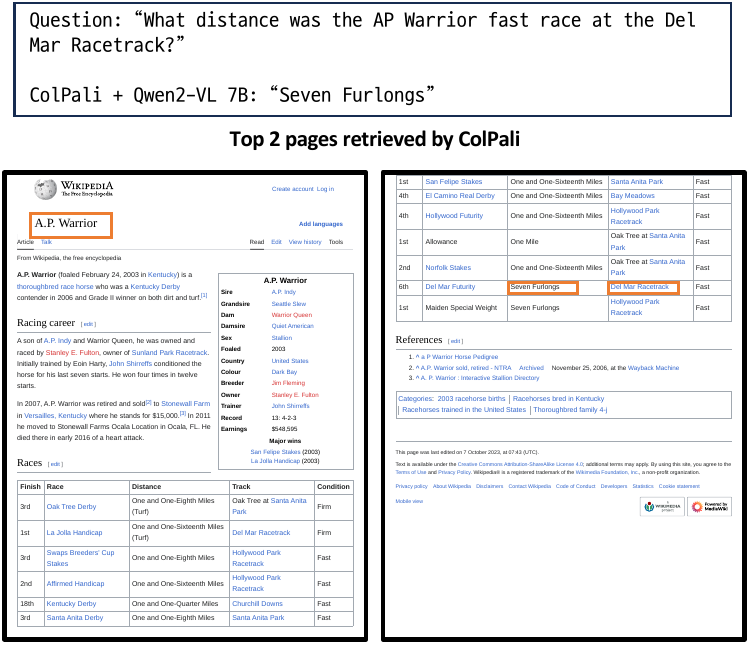}
    \caption{Qualitative example of \colpali{} + Qwen2-VL 7B on \dataset{}. Image regions relevant to the question/answer are highlighted with orange boxes. The question requires multi-page/document reasoning.
    }
    \label{fig:qual_example_apwarrior}
\end{figure*}

\begin{figure}[h]
    \centering
    \includegraphics[width=\linewidth]{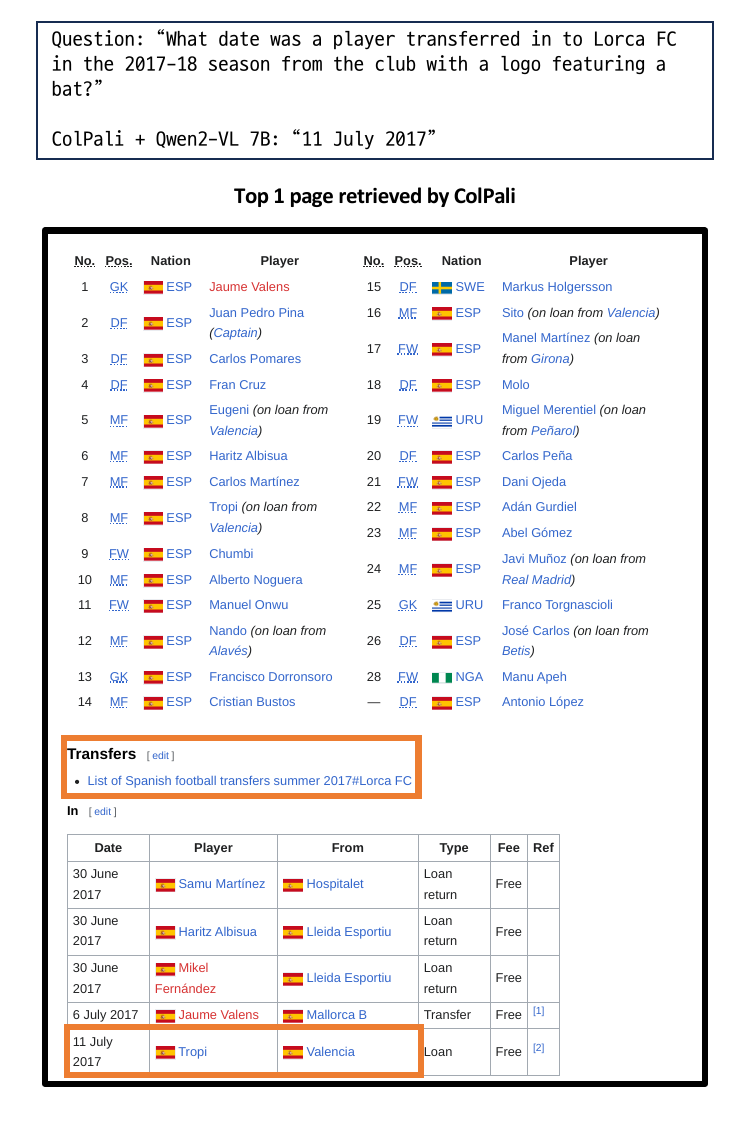}
    \caption{Qualitative example of \colpali{} + Qwen2-VL 7B on \dataset{}. Image regions relevant to the question/answer are highlighted with orange boxes.
    The VQA component could combine both the retrieved knowledge (Tropi was transferred on 11 July 2017) and its own knowledge (Valencia CF has a logo with a bat) to provide the final answer.
    }
    \label{fig:qual_example_valencia}
\end{figure}

\subsection{Qualitative Examples}
\label{sec:qual_examples}

In \cref{fig:qual_example_daysgone}, \cref{fig:qual_example_apwarrior}, and \cref{fig:qual_example_valencia},
we provide qualitative examples of \method{} (\colpali{} + Qwen2-VL 7B)'s question answering results on several \dataset{} examples.
In \cref{fig:qual_example_daysgone},
the answer information is only visually stored within the game logo (`man is leaning on a motorcycle'), and \method{} could find the information.
In \cref{fig:qual_example_apwarrior},
the question requires multi-hop reasoning across different pages/documents, and \method{} could combine information from multiple retrieved pages.
In \cref{fig:qual_example_valencia}, although \colpali{} did not retrieve the page that contains information about a team whose logo features a bat,
Qwen-2 VL leverages its own knowledge `Valencia CF has a logo featuring a bat',
and could provide the final answer.
Overall, the qualitative examples showcase that \method{} can successfully tackle different questions whose answer sources exist in various modalities.

\section{Related Work}
\label{sec:related_work}

\paragraph{Document visual question answering.}
\citet{mathew2021docvqa} proposed document visual question answering (DocVQA) task,
where a model extracts information from documents by treating them as images, like in generic visual question answering~\citep{Antol2015}.
Most research on DocVQA focuses on handling a single-page document~\citep{mathew2021docvqa,Mathew2021InfographicVQA,hu2024mplugdocowl,Wang2023docllm,tang2023unifying,LayoutLMv3,kim2022donut,Pix2Struct,Ye2023UReaderUO},
and it has been now a common practice to include the single-page DocVQA~\citep{mathew2021docvqa} as a part of the image understanding evaluation suite among recent MLMs~\citep{chen2024internvl2,Qwen2VL,Beyer2024paligemma,laurençon2024idefics3,gpt4o,geminiteam2024gemini}.
Several recent works study applying MLMs for DocVQA on multi-page documents~\citep{Landeghem2023dude,tito2023mpdocvqa,ma2024mmlongbench}.
However, all previous works on DocVQA have focused on handling questions in the context of a specific document, such as ``What was the gross profit in the year
2009?''~\citep{mathew2021docvqa,tito2023mpdocvqa,ding2023pdfvqa,ma2024mmlongbench}.
While this is probably due to the limited context length of the backbone multi-modal LMs,
this does not reflect real-world scenarios, where users often ask questions that require information across different pages/documents.
We address the limitation and propose \method{} framework and \dataset{} dataset for effective, efficient, and flexible document understanding under
various document contexts (closed-domain and open-domain),
question hops (single-hop and multi-hop),
and evidence modalities (text, chart, figure, \etc{}).

\paragraph{Retrieval-augmented generation.}
Retrieval-augmented generation (RAG)~\citep{Lewis2020} has emerged as a hybrid approach combining retrieval systems with generative models to improve the quality and relevance of generated content~\citep{gao2023retrieval}.
RAG has been widely studied for open-domain question answering~\citep{Guu2020,zhu2021retrieving,Karpukhin2020DPR,Izacard2021,luo2023sailsearchaugmentedinstructionlearning,asai2023selfraglearningretrievegenerate}, where the community has well-established practices for text-based pipelines.
A line of work in VQA
studies RAG on visual questions that require world knowledge
\citep{Yasunaga2022,chen2022murag,Mensink2023Encyclopedic,schwenk2022aokvqabenchmarkvisualquestion},
but their retrieval context is usually generic images and/or short text snippets and does not cover DocVQA settings.
To the best of our knowledge, no prior work has explored RAG setting for multi-modal document understanding only with multi-modal models (instead of using OCR methods).
Our framework tackles open-domain question answering over documents with complex multi-modal contexts, including textual, tabular, and visual information across different pages and documents.

\section{Conclusion}
\label{sec:conclusion}

We introduce \method{},
a novel multi-modal RAG framework that
flexibly accommodates
various document contexts (closed-domain and open-domain),
question hops (single-hop and multi-hop),
and evidence modalities (text, chart, figure, \etc{}).
In \method{},
a multi-modal retrieval model identifies relevant pages from single or multiple documents, which are then processed by a multi-modal language model, where all documents are represented as pixels. 
Next, we introduce \dataset{}, the first benchmark that evaluates open-domain multi-modal document understanding capabilities. \dataset{} consists of 2,000+ questions and 3,000+ PDF documents, and the questions need to be answered with various modalities such as images, text, and tables.
Our experiments in three datasets (\dataset{}, MP-DocVQA, and MMLongBench-Doc) demonstrate significant advantages of \method{} over existing methods, including the state-of-the-art performance in MP-DocVQA.
We also provide analysis comparing different indexing strategies, multi-modal LMs, and multi-modal retrieval models.
Finally, we show qualitative examples where \method{} can successfully tackle different questions whose answer sources exist in various modalities.
We hope that our work encourages future advancements in multi-modal frameworks for document understanding, paving the way for more robust, scalable, and practical solutions in real-world applications.

\section*{Ethical Considerations}
\label{sec:ethics}

\paragraph{Limitations.}
Since our multimodal retrieval models and multimodal LMs were trained with English-heavy datasets, they might not understand prompts or documents written in non-English.
While our \method{} framework can benefit many document understanding applications, the model components could present false or biased information. Thus, the framework should be used with human supervision in real-world applications. Note that \method{} is designed with flexibility so that users can update or replace components as more accurate solutions for each element of the framework become available in the future.

\paragraph{Data collection.}
We do not involve human subjects during data collection.
We do not claim ownership/rights of the Wikipedia documents, and we attribute the source Wikipedia document URLs to all pages.

{
    \small
    \bibliographystyle{ieeenat_fullname}
    \bibliography{references}

\begin{thebibliography}{66}
\providecommand{\natexlab}[1]{#1}
\providecommand{\url}[1]{\texttt{#1}}
\expandafter\ifx\csname urlstyle\endcsname\relax
  \providecommand{\doi}[1]{doi: #1}\else
  \providecommand{\doi}{doi: \begingroup \urlstyle{rm}\Url}\fi

\bibitem[Antol et~al.(2015)Antol, Agrawal, Lu, Mitchell, Batra, Zitnick, and Parikh]{Antol2015}
Stanislaw Antol, Aishwarya Agrawal, Jiasen Lu, Margaret Mitchell, Dhruv Batra, C.~Lawrence Zitnick, and Devi Parikh.
\newblock {VQA: Visual question answering}.
\newblock In \emph{ICCV}, 2015.

\bibitem[Asai et~al.(2023)Asai, Wu, Wang, Sil, and Hajishirzi]{asai2023selfraglearningretrievegenerate}
Akari Asai, Zeqiu Wu, Yizhong Wang, Avirup Sil, and Hannaneh Hajishirzi.
\newblock Self-rag: Learning to retrieve, generate, and critique through self-reflection, 2023.

\bibitem[Baechler et~al.(2024)Baechler, Sunkara, Wang, Zubach, Mansoor, Etter, Cărbune, Lin, Chen, and Sharma]{Baechler2024screenai}
Gilles Baechler, Srinivas Sunkara, Maria Wang, Fedir Zubach, Hassan Mansoor, Vincent Etter, Victor Cărbune, Jason Lin, Jindong Chen, and Abhanshu Sharma.
\newblock {ScreenAI: A Vision-Language Model for UI and Infographics Understanding}, 2024.

\bibitem[Bai et~al.(2023)Bai, Bai, Yang, Wang, Tan, Wang, Lin, Zhou, and Zhou]{bai2023qwenvl}
Jinze Bai, Shuai Bai, Shusheng Yang, Shijie Wang, Sinan Tan, Peng Wang, Junyang Lin, Chang Zhou, and Jingren Zhou.
\newblock {Qwen-VL}: A frontier large vision-language model with versatile abilities.
\newblock \emph{arXiv preprint}, abs/2308.12966, 2023.

\bibitem[Bai et~al.(2024)Bai, Lv, Zhang, He, Qi, Hou, Tang, Dong, and Li]{bai2024longalign}
Yushi Bai, Xin Lv, Jiajie Zhang, Yuze He, Ji Qi, Lei Hou, Jie Tang, Yuxiao Dong, and Juanzi Li.
\newblock {LongAlign}: A recipe for long context alignment of large language models.
\newblock \emph{arXiv preprint}, abs/2401.18058, 2024.

\bibitem[Belval(2017)]{pdf2image}
Edouard Belval.
\newblock pdf2image, 2017.

\bibitem[Beyer et~al.(2024)Beyer, Steiner, Pinto, Kolesnikov, Wang, Salz, Neumann, Alabdulmohsin, Tschannen, Bugliarello, Unterthiner, Keysers, Koppula, Liu, Grycner, Gritsenko, Houlsby, Kumar, Rong, Eisenschlos, Kabra, Bauer, Bo{\v{s}}njak, Chen, Minderer, Voigtlaender, Bica, Balazevic, Puigcerver, Papalampidi, Henaff, Xiong, Soricut, Harmsen, and Zhai]{Beyer2024paligemma}
Lucas Beyer, Andreas Steiner, Andr{\'{e}}~Susano Pinto, Alexander Kolesnikov, Xiao Wang, Daniel Salz, Maxim Neumann, Ibrahim Alabdulmohsin, Michael Tschannen, Emanuele Bugliarello, Thomas Unterthiner, Daniel Keysers, Skanda Koppula, Fangyu Liu, Adam Grycner, Alexey Gritsenko, Neil Houlsby, Manoj Kumar, Keran Rong, Julian Eisenschlos, Rishabh Kabra, Matthias Bauer, Matko Bo{\v{s}}njak, Xi Chen, Matthias Minderer, Paul Voigtlaender, Ioana Bica, Ivana Balazevic, Joan Puigcerver, Pinelopi Papalampidi, Olivier Henaff, Xi Xiong, Radu Soricut, Jeremiah Harmsen, and Xiaohua Zhai.
\newblock {PaliGemma: A versatile 3B VLM for transfer}, 2024.

\bibitem[Biten et~al.(2019)Biten, Mafla, Gomez, Jawahar, and Karatzas]{Biten2019}
Ali~Furkan Biten, Andres Mafla, Lluis Gomez, Valveny C~V Jawahar, and Dimosthenis Karatzas.
\newblock {Scene Text Visual Question Answering}.
\newblock In \emph{ICCV}, 2019.

\bibitem[Blau et~al.(2024)Blau, Fogel, Ronen, Golts, Ganz, Ben~Avraham, Aberdam, Tsiper, and Litman]{blau2024gram}
Tsachi Blau, Sharon Fogel, Roi Ronen, Alona Golts, Roy Ganz, Elad Ben~Avraham, Aviad Aberdam, Shahar Tsiper, and Ron Litman.
\newblock Gram: Global reasoning for multi-page vqa.
\newblock In \emph{Proceedings of the IEEE/CVF Conference on Computer Vision and Pattern Recognition}, pages 15598--15607, 2024.

\bibitem[Borchmann et~al.(2024)Borchmann, Pietruszka, Ja{\'{s}}kowski, Jurkiewicz, Halama, J{\'{o}}ziak, Garncarek, Liskowski, Szyndler, Gretkowski, O{\l}tusek, Nowakowska, Zaw{\l}ocki, Duhr, Dyda, and Turski]{Borchmann2024arctictilt}
{\L}ukasz Borchmann, Micha{\l} Pietruszka, Wojciech Ja{\'{s}}kowski, Dawid Jurkiewicz, Piotr Halama, Pawe{\l} J{\'{o}}ziak, {\L}ukasz Garncarek, Pawe{\l} Liskowski, Karolina Szyndler, Andrzej Gretkowski, Julita O{\l}tusek, Gabriela Nowakowska, Artur Zaw{\l}ocki, {\L}ukasz Duhr, Pawe{\l} Dyda, and Micha{\l} Turski.
\newblock {Arctic-TILT. Business Document Understanding at Sub-Billion Scale}, 2024.

\bibitem[Chen et~al.(2022)Chen, Hu, Chen, Verga, and Cohen]{chen2022murag}
Wenhu Chen, Hexiang Hu, Xi Chen, Pat Verga, and William~W Cohen.
\newblock Murag: Multimodal retrieval-augmented generator for open question answering over images and text.
\newblock \emph{arXiv preprint arXiv:2210.02928}, 2022.

\bibitem[Chen et~al.(2024)Chen, Wang, Tian, Ye, Gao, Cui, Tong, Hu, Luo, Ma, et~al.]{chen2024internvl2}
Zhe Chen, Weiyun Wang, Hao Tian, Shenglong Ye, Zhangwei Gao, Erfei Cui, Wenwen Tong, Kongzhi Hu, Jiapeng Luo, Zheng Ma, et~al.
\newblock How far are we to gpt-4v? closing the gap to commercial multimodal models with open-source suites.
\newblock \emph{arXiv preprint arXiv:2404.16821}, 2024.

\bibitem[Dao(2024)]{dao2023flashattention2}
Tri Dao.
\newblock Flash{A}ttention-2: Faster attention with better parallelism and work partitioning.
\newblock In \emph{International Conference on Learning Representations (ICLR)}, 2024.

\bibitem[Ding et~al.(2023)Ding, Luo, Chung, and Han]{ding2023pdfvqa}
Yihao Ding, Siwen Luo, Hyunsuk Chung, and Soyeon~Caren Han.
\newblock Pdfvqa: A new dataset for real-world vqa on pdf documents.
\newblock In \emph{Joint European Conference on Machine Learning and Knowledge Discovery in Databases}, pages 585--601. Springer, 2023.

\bibitem[Dong et~al.(2024)Dong, Zhang, Zang, Cao, Wang, Ouyang, Zhang, Duan, Zhang, Li, et~al.]{dong2024internlm}
Xiaoyi Dong, Pan Zhang, Yuhang Zang, Yuhang Cao, Bin Wang, Linke Ouyang, Songyang Zhang, Haodong Duan, Wenwei Zhang, Yining Li, et~al.
\newblock {Internlm-Xcomposer2-4KHD}: A pioneering large vision-language model handling resolutions from 336 pixels to 4k hd.
\newblock \emph{arXiv preprint}, abs/2404.06512, 2024.

\bibitem[Douze et~al.(2024)Douze, Guzhva, Deng, Johnson, Szilvasy, Mazaré, Lomeli, Hosseini, and Jégou]{douze2024faiss}
Matthijs Douze, Alexandr Guzhva, Chengqi Deng, Jeff Johnson, Gergely Szilvasy, Pierre-Emmanuel Mazaré, Maria Lomeli, Lucas Hosseini, and Hervé Jégou.
\newblock The faiss library, 2024.

\bibitem[Faysse et~al.(2024)Faysse, Sibille, Wu, Omrani, Viaud, Hudelot, and Colombo]{Faysse2024colpali}
Manuel Faysse, Hugues Sibille, Tony Wu, Bilel Omrani, Gautier Viaud, C{\'{e}}line Hudelot, and Pierre Colombo.
\newblock {ColPali: Efficient Document Retrieval with Vision Language Models}, 2024.

\bibitem[Fenniak and {PyPDF2 Contributors}(2022)]{pypdf2}
Mathieu Fenniak and {PyPDF2 Contributors}.
\newblock The {PyPDF2} library, version 2, 2022.

\bibitem[Gao et~al.(2023)Gao, Xiong, Gao, Jia, Pan, Bi, Dai, Sun, and Wang]{gao2023retrieval}
Yunfan Gao, Yun Xiong, Xinyu Gao, Kangxiang Jia, Jinliu Pan, Yuxi Bi, Yi Dai, Jiawei Sun, and Haofen Wang.
\newblock Retrieval-augmented generation for large language models: A survey.
\newblock \emph{arXiv preprint arXiv:2312.10997}, 2023.

\bibitem[{Gemini Team}(2024)]{geminiteam2024gemini}
{Gemini Team}.
\newblock {Gemini 1.5}: Unlocking multimodal understanding across millions of tokens of context, 2024.

\bibitem[Guu et~al.(2020)Guu, Lee, Tung, Pasupat, and Chang]{Guu2020}
Kelvin Guu, Kenton Lee, Zora Tung, Panupong Pasupat, and Ming-Wei Chang.
\newblock {REALM: Retrieval-Augmented Language Model Pre-Training}.
\newblock In \emph{ICML}, 2020.

\bibitem[Hu et~al.(2024)Hu, Xu, Ye, Yan, Zhang, Zhang, Li, Zhang, Jin, Huang, and Zhou]{hu2024mplugdocowl}
Anwen Hu, Haiyang Xu, Jiabo Ye, Ming Yan, Liang Zhang, Bo Zhang, Chen Li, Ji Zhang, Qin Jin, Fei Huang, and Jingren Zhou.
\newblock mplug-docowl 1.5: Unified structure learning for ocr-free document understanding, 2024.

\bibitem[Huang et~al.(2022)Huang, Lv, Cui, Lu, and Wei]{LayoutLMv3}
Yupan Huang, Tengchao Lv, Lei Cui, Yutong Lu, and Furu Wei.
\newblock Layoutlmv3: Pre-training for document ai with unified text and image masking.
\newblock In \emph{Proceedings of the 30th ACM International Conference on Multimedia}, page 4083–4091, New York, NY, USA, 2022. Association for Computing Machinery.

\bibitem[Izacard and Grave(2021)]{Izacard2021}
Gautier Izacard and Edouard Grave.
\newblock {Leveraging Passage Retrieval with Generative Models for Open Domain Question Answering}.
\newblock In \emph{EACL}, 2021.

\bibitem[Jiang et~al.(2023)Jiang, Sablayrolles, Mensch, Bamford, Chaplot, de~las Casas, Bressand, Lengyel, Lample, Saulnier, Lavaud, Lachaux, Stock, Scao, Lavril, Wang, Lacroix, and Sayed]{jiang2023mistral}
Albert~Q. Jiang, Alexandre Sablayrolles, Arthur Mensch, Chris Bamford, Devendra~Singh Chaplot, Diego de~las Casas, Florian Bressand, Gianna Lengyel, Guillaume Lample, Lucile Saulnier, Lélio~Renard Lavaud, Marie-Anne Lachaux, Pierre Stock, Teven~Le Scao, Thibaut Lavril, Thomas Wang, Timothée Lacroix, and William~El Sayed.
\newblock Mistral 7b, 2023.

\bibitem[Johnson et~al.(2021)Johnson, Douze, and Jégou]{faiss2021}
Jeff Johnson, Matthijs Douze, and Hervé Jégou.
\newblock Billion-scale similarity search with gpus.
\newblock \emph{IEEE Transactions on Big Data}, 7\penalty0 (3):\penalty0 535--547, 2021.

\bibitem[Jégou et~al.(2011)Jégou, Douze, and Schmid]{productquantization}
Herve Jégou, Matthijs Douze, and Cordelia Schmid.
\newblock Product quantization for nearest neighbor search.
\newblock \emph{IEEE Transactions on Pattern Analysis and Machine Intelligence}, 33\penalty0 (1):\penalty0 117--128, 2011.

\bibitem[Karpukhin et~al.(2020)Karpukhin, Barlas, Min, Lewis, Wu, Edunov, Chen, and Yih]{Karpukhin2020DPR}
Vladimir Karpukhin, O Barlas, Sewon Min, Patrick Lewis, Ledell Wu, Sergey Edunov, Danqi Chen, and Wen-tau Yih.
\newblock {Dense Passage Retrieval for Open-Domain Question Answering}.
\newblock In \emph{EMNLP}, pages 6769--6781, 2020.

\bibitem[Khattab and Zaharia(2020)]{Khattab2020colbert}
Omar Khattab and Matei Zaharia.
\newblock {ColBERT: Efficient and Effective Passage Search via Contextualized Late Interaction over BERT}.
\newblock \emph{SIGIR 2020 - Proceedings of the 43rd International ACM SIGIR Conference on Research and Development in Information Retrieval}, pages 39--48, 2020.

\bibitem[Kim et~al.(2022)Kim, Hong, Yim, Nam, Park, Yim, Hwang, Yun, Han, and Park]{kim2022donut}
Geewook Kim, Teakgyu Hong, Moonbin Yim, JeongYeon Nam, Jinyoung Park, Jinyeong Yim, Wonseok Hwang, Sangdoo Yun, Dongyoon Han, and Seunghyun Park.
\newblock Ocr-free document understanding transformer.
\newblock In \emph{European Conference on Computer Vision (ECCV)}, 2022.

\bibitem[Landeghem et~al.(2023)Landeghem, Powalski, Tito, Jurkiewicz, Blaschko, Borchmann, Coustaty, Moens, Pietruszka, Ackaert, Stanis{\l}awek, J{\'{o}}ziak, and Valveny]{Landeghem2023dude}
Jordy~Van Landeghem, Rafa{\l} Powalski, Rub{\`{e}}n Tito, Dawid Jurkiewicz, Matthew Blaschko, {\L}ukasz Borchmann, Micka{\"{e}}l Coustaty, Sien Moens, Micha{\l} Pietruszka, Bertrand Ackaert, Tomasz Stanis{\l}awek, Pawe{\l} J{\'{o}}ziak, and Ernest Valveny.
\newblock {Document Understanding Dataset and Evaluation (DUDE)}.
\newblock In \emph{ICCV}, 2023.

\bibitem[Laurençon et~al.(2024{\natexlab{a}})Laurençon, Marafioti, Sanh, and Tronchon]{laurençon2024idefics3}
Hugo Laurençon, Andrés Marafioti, Victor Sanh, and Léo Tronchon.
\newblock Building and better understanding vision-language models: insights and future directions, 2024{\natexlab{a}}.

\bibitem[Laurençon et~al.(2024{\natexlab{b}})Laurençon, Tronchon, Cord, and Sanh]{laurencon2024mattersidefics2}
Hugo Laurençon, Léo Tronchon, Matthieu Cord, and Victor Sanh.
\newblock What matters when building vision-language models?, 2024{\natexlab{b}}.

\bibitem[Lee et~al.(2023)Lee, Joshi, Turc, Hu, Liu, Eisenschlos, Khandelwal, Shaw, Chang, and Toutanova]{Pix2Struct}
Kenton Lee, Mandar Joshi, Iulia Turc, Hexiang Hu, Fangyu Liu, Julian Eisenschlos, Urvashi Khandelwal, Peter Shaw, Ming-Wei Chang, and Kristina Toutanova.
\newblock Pix2struct: screenshot parsing as pretraining for visual language understanding.
\newblock In \emph{Proceedings of the 40th International Conference on Machine Learning}. JMLR.org, 2023.

\bibitem[Lewis et~al.(2020)Lewis, Perez, Piktus, Petroni, Karpukhin, Goyal, K{\"{u}}ttler, Lewis, Yih, Rockt{\"{a}}schel, Riedel, and Kiela]{Lewis2020}
Patrick Lewis, Ethan Perez, Aleksandra Piktus, Fabio Petroni, Vladimir Karpukhin, Naman Goyal, Heinrich K{\"{u}}ttler, Mike Lewis, Wen~Tau Yih, Tim Rockt{\"{a}}schel, Sebastian Riedel, and Douwe Kiela.
\newblock {Retrieval-augmented generation for knowledge-intensive NLP tasks}.
\newblock In \emph{NeurIPS}, 2020.

\bibitem[Li et~al.(2023)Li, Yang, Liu, Ma, Zhang, Yang, Sun, Liu, and Bai]{li2023monkey}
Zhang Li, Biao Yang, Qiang Liu, Zhiyin Ma, Shuo Zhang, Jingxu Yang, Yabo Sun, Yuliang Liu, and Xiang Bai.
\newblock {Monkey}: Image resolution and text label are important things for large multi-modal models.
\newblock \emph{arXiv preprint}, abs/2311.06607, 2023.

\bibitem[{Llama Team}(2024)]{dubey2024llama3herdmodels}
{Llama Team}.
\newblock The llama 3 herd of models, 2024.

\bibitem[Lu et~al.(2024)Lu, Liu, Zhang, Wang, Dong, Liu, Sun, Ren, Li, Sun, et~al.]{lu2024deepseek}
Haoyu Lu, Wen Liu, Bo Zhang, Bingxuan Wang, Kai Dong, Bo Liu, Jingxiang Sun, Tongzheng Ren, Zhuoshu Li, Yaofeng Sun, et~al.
\newblock {DeepSeek-VL}: towards real-world vision-language understanding.
\newblock \emph{arXiv preprint}, abs/2403.05525, 2024.

\bibitem[Luo et~al.(2023)Luo, Chuang, Gong, Zhang, Kim, Wu, Fox, Meng, and Glass]{luo2023sailsearchaugmentedinstructionlearning}
Hongyin Luo, Yung-Sung Chuang, Yuan Gong, Tianhua Zhang, Yoon Kim, Xixin Wu, Danny Fox, Helen Meng, and James Glass.
\newblock Sail: Search-augmented instruction learning, 2023.

\bibitem[Ma et~al.(2024)Ma, Zang, Chen, Chen, Jiao, Li, Lu, Liu, Ma, Dong, et~al.]{ma2024mmlongbench}
Yubo Ma, Yuhang Zang, Liangyu Chen, Meiqi Chen, Yizhu Jiao, Xinze Li, Xinyuan Lu, Ziyu Liu, Yan Ma, Xiaoyi Dong, et~al.
\newblock Mmlongbench-doc: Benchmarking long-context document understanding with visualizations.
\newblock \emph{arXiv preprint arXiv:2407.01523}, 2024.

\bibitem[Mathew et~al.(2021{\natexlab{a}})Mathew, Bagal, Tito, Karatzas, Valveny, and Jawahar]{Mathew2021InfographicVQA}
Minesh Mathew, Viraj Bagal, Rub{\`e}n~P{\'e}rez Tito, Dimosthenis Karatzas, Ernest Valveny, and C.V. Jawahar.
\newblock Infographicvqa.
\newblock \emph{2022 IEEE/CVF Winter Conference on Applications of Computer Vision (WACV)}, pages 2582--2591, 2021{\natexlab{a}}.

\bibitem[Mathew et~al.(2021{\natexlab{b}})Mathew, Karatzas, and Jawahar]{mathew2021docvqa}
Minesh Mathew, Dimosthenis Karatzas, and CV Jawahar.
\newblock Docvqa: A dataset for vqa on document images.
\newblock In \emph{Proceedings of the IEEE/CVF winter conference on applications of computer vision}, pages 2200--2209, 2021{\natexlab{b}}.

\bibitem[Memon et~al.(2020)Memon, Sami, Khan, and Uddin]{review_ocr}
Jamshed Memon, Maira Sami, Rizwan~Ahmed Khan, and Mueen Uddin.
\newblock Handwritten optical character recognition (ocr): A comprehensive systematic literature review (slr).
\newblock \emph{IEEE Access}, 8:\penalty0 142642--142668, 2020.

\bibitem[Mensink et~al.(2023)Mensink, Uijlings, Castrejon, Goel, Cadar, Zhou, Sha, Araujo, and Ferrari]{Mensink2023Encyclopedic}
Thomas Mensink, Jasper Uijlings, Lluis Castrejon, Arushi Goel, Felipe Cadar, Howard Zhou, Fei Sha, Andr{\'{e}} Araujo, and Vittorio Ferrari.
\newblock {Encyclopedic VQA: Visual questions about detailed properties of fine-grained categories}.
\newblock In \emph{Proceedings of the IEEE International Conference on Computer Vision}, pages 3090--3101, 2023.

\bibitem[Microsoft(2021)]{playwright}
Microsoft.
\newblock Playwright for python, 2021.

\bibitem[{OpenAI}(2024)]{gpt4o}
{OpenAI}.
\newblock Hello gpt-4o, 2024.

\bibitem[Paszke et~al.(2017)Paszke, Gross, Chintala, Chana, Yang, DeVito, Lin, Desmaison, Antiga, and Lerer]{Paszke2017}
Adam Paszke, Sam Gross, Soumith Chintala, Gregory Chana, Edward Yang, Zachary DeVito, Zeming Lin, Alban Desmaison, Luca Antiga, and Adam Lerer.
\newblock {Automatic differentiation in PyTorch}.
\newblock In \emph{NIPS Workshop}, 2017.

\bibitem[Paszke et~al.(2019)Paszke, Gross, Massa, Lerer, Bradbury, Chanan, Killeen, Lin, Gimelshein, Antiga, Desmaison, K{\"{o}}pf, Yang, DeVito, Raison, Tejani, Chilamkurthy, Steiner, Fang, Bai, and Chintala]{Paszke2019}
Adam Paszke, Sam Gross, Francisco Massa, Adam Lerer, James Bradbury, Gregory Chanan, Trevor Killeen, Zeming Lin, Natalia Gimelshein, Luca Antiga, Alban Desmaison, Andreas K{\"{o}}pf, Edward Yang, Zach DeVito, Martin Raison, Alykhan Tejani, Sasank Chilamkurthy, Benoit Steiner, Lu Fang, Junjie Bai, and Soumith Chintala.
\newblock {PyTorch: An imperative style, high-performance deep learning library}.
\newblock \emph{Advances in Neural Information Processing Systems}, 32\penalty0 (NeurIPS), 2019.

\bibitem[pdfminer(2019)]{pdfminer}
pdfminer.
\newblock pdfminer.six, 2019.

\bibitem[Santhanam et~al.(2022)Santhanam, Khattab, Saad-Falcon, Potts, and Zaharia]{Santhanam2022colbert2}
Keshav Santhanam, Omar Khattab, Jon Saad-Falcon, Christopher Potts, and Matei Zaharia.
\newblock {ColBERTv2: Effective and Efficient Retrieval via Lightweight Late Interaction}.
\newblock \emph{NAACL 2022 - 2022 Conference of the North American Chapter of the Association for Computational Linguistics: Human Language Technologies, Proceedings of the Conference}, pages 3715--3734, 2022.

\bibitem[Schwenk et~al.(2022)Schwenk, Khandelwal, Clark, Marino, and Mottaghi]{schwenk2022aokvqabenchmarkvisualquestion}
Dustin Schwenk, Apoorv Khandelwal, Christopher Clark, Kenneth Marino, and Roozbeh Mottaghi.
\newblock A-okvqa: A benchmark for visual question answering using world knowledge, 2022.

\bibitem[Sivic and Zisserman(2003)]{invertedvideogoogle}
Sivic and Zisserman.
\newblock Video google: a text retrieval approach to object matching in videos.
\newblock In \emph{Proceedings Ninth IEEE International Conference on Computer Vision}, pages 1470--1477 vol.2, 2003.

\bibitem[Smith(2007)]{smith2007overview}
Ray Smith.
\newblock An overview of the tesseract ocr engine.
\newblock In \emph{ICDAR}, 2007.

\bibitem[Talmor et~al.(2021)Talmor, Yoran, Catav, Lahav, Wang, Asai, Ilharco, Hajishirzi, and Berant]{talmor2021multimodalqa}
Alon Talmor, Ori Yoran, Amnon Catav, Dan Lahav, Yizhong Wang, Akari Asai, Gabriel Ilharco, Hannaneh Hajishirzi, and Jonathan Berant.
\newblock Multimodalqa: Complex question answering over text, tables and images.
\newblock \emph{arXiv preprint arXiv:2104.06039}, 2021.

\bibitem[Tang et~al.(2023)Tang, Yang, Wang, Fang, Liu, Zhu, Zeng, Zhang, and Bansal]{tang2023unifying}
Zineng Tang, Ziyi Yang, Guoxin Wang, Yuwei Fang, Yang Liu, Chenguang Zhu, Michael Zeng, Cha Zhang, and Mohit Bansal.
\newblock Unifying vision, text, and layout for universal document processing, 2023.

\bibitem[{The Chromium Project Authors}(2024)]{chromium}
{The Chromium Project Authors}.
\newblock The chromium projects, 2024.

\bibitem[Tito et~al.(2023)Tito, Karatzas, and Valveny]{tito2023mpdocvqa}
Rub{\`e}n Tito, Dimosthenis Karatzas, and Ernest Valveny.
\newblock Hierarchical multimodal transformers for multipage docvqa.
\newblock \emph{Pattern Recognition}, 144:\penalty0 109834, 2023.

\bibitem[Wang et~al.(2023)Wang, Raman, Sibue, Ma, Babkin, Kaur, Pei, Nourbakhsh, and Liu]{Wang2023docllm}
Dongsheng Wang, Natraj Raman, Mathieu Sibue, Zhiqiang Ma, Petr Babkin, Simerjot Kaur, Yulong Pei, Armineh Nourbakhsh, and Xiaomo Liu.
\newblock {DocLLM: A layout-aware generative language model for multimodal document understanding}, 2023.

\bibitem[Wang et~al.(2024)Wang, Bai, Tan, Wang, Fan, Bai, Chen, Liu, Wang, Ge, Fan, Dang, Du, Ren, Men, Liu, Zhou, Zhou, and Lin]{Qwen2VL}
Peng Wang, Shuai Bai, Sinan Tan, Shijie Wang, Zhihao Fan, Jinze Bai, Keqin Chen, Xuejing Liu, Jialin Wang, Wenbin Ge, Yang Fan, Kai Dang, Mengfei Du, Xuancheng Ren, Rui Men, Dayiheng Liu, Chang Zhou, Jingren Zhou, and Junyang Lin.
\newblock Qwen2-vl: Enhancing vision-language model's perception of the world at any resolution.
\newblock \emph{arXiv preprint arXiv:2409.12191}, 2024.

\bibitem[Wolf et~al.(2020)Wolf, Debut, Sanh, Chaumond, Delangue, Moi, Cistac, Rault, Louf, Funtowicz, and Brew]{Wolf2019}
Thomas Wolf, Lysandre Debut, Victor Sanh, Julien Chaumond, Clement Delangue, Anthony Moi, Pierric Cistac, Tim Rault, R{\'{e}}mi Louf, Morgan Funtowicz, and Jamie Brew.
\newblock {HuggingFace's Transformers: State-of-the-art Natural Language Processing}.
\newblock In \emph{EMNLP}, 2020.

\bibitem[Xu et~al.(2024)Xu, Yao, Guo, Cui, Ni, Ge, Chua, Liu, and Huang]{xu2024llava-uhd}
Ruyi Xu, Yuan Yao, Zonghao Guo, Junbo Cui, Zanlin Ni, Chunjiang Ge, Tat-Seng Chua, Zhiyuan Liu, and Gao Huang.
\newblock {LLaVA-UHD}: An {LMM} perceiving any aspect ratio and high-resolution images.
\newblock \emph{arXiv preprint}, abs/2403.11703, 2024.

\bibitem[Yasunaga et~al.(2023)Yasunaga, Aghajanyan, Shi, James, Leskovec, Liang, Lewis, Zettlemoyer, and Yih]{Yasunaga2022}
Michihiro Yasunaga, Armen Aghajanyan, Weijia Shi, Rich James, Jure Leskovec, Percy Liang, Mike Lewis, Luke Zettlemoyer, and Wen-tau Yih.
\newblock {Retrieval-Augmented Multimodal Language Modeling}.
\newblock In \emph{ICML}, 2023.

\bibitem[Ye et~al.(2023)Ye, Hu, Xu, Ye, Yan, Xu, Li, Tian, Qian, Zhang, Jin, He, Lin, and Huang]{Ye2023UReaderUO}
Jiabo Ye, Anwen Hu, Haiyang Xu, Qinghao Ye, Mingshi Yan, Guohai Xu, Chenliang Li, Junfeng Tian, Qi Qian, Ji Zhang, Qin Jin, Liang He, Xin Lin, and Feiyan Huang.
\newblock Ureader: Universal ocr-free visually-situated language understanding with multimodal large language model.
\newblock In \emph{Conference on Empirical Methods in Natural Language Processing}, 2023.

\bibitem[Yu et~al.(2024)Yu, Zhang, Yao, Dang, Chen, Lu, Cui, He, Liu, Chua, and Sun]{yu2024rlaifv}
Tianyu Yu, Haoye Zhang, Yuan Yao, Yunkai Dang, Da Chen, Xiaoman Lu, Ganqu Cui, Taiwen He, Zhiyuan Liu, Tat-Seng Chua, and Maosong Sun.
\newblock {RLAIF-V}: Aligning mllms through open-source ai feedback for super gpt-4v trustworthiness.
\newblock \emph{arXiv preprint}, abs/2405.17220, 2024.

\bibitem[Zhu et~al.(2021)Zhu, Lei, Wang, Zheng, Poria, and Chua]{zhu2021retrieving}
Fengbin Zhu, Wenqiang Lei, Chao Wang, Jianming Zheng, Soujanya Poria, and Tat-Seng Chua.
\newblock Retrieving and reading: A comprehensive survey on open-domain question answering.
\newblock \emph{arXiv preprint arXiv:2101.00774}, 2021.

\bibitem[Zobel and Moffat(2006)]{invertedfiles}
Justin Zobel and Alistair Moffat.
\newblock Inverted files for text search engines.
\newblock \emph{ACM Comput. Surv.}, 38\penalty0 (2):\penalty0 6–es, 2006.

\end{thebibliography}
}

\end{document}